# *MedCAT* - MEDICAL CONCEPT ANNOTATION TOOL

**Zeljko Kraljevic[1], Daniel Bean[1,2], Aurelie Mascio[1], Lukasz Roguski[2,3], Amos Folarin[1,3,4], Angus Roberts[1,2,4], Rebecca Bendayan[1,4], Richard Dobson[1,2,3,4]**
{firstname.lastname}@kcl.ac.uk

Affiliations
1. Department of Biostatistics and Health Informatics, Institute of Psychiatry, Psychology and Neuroscience, King's College London, London, U.K.
2. Health Data Research UK London, University College London, London, U.K.
3. Institute of Health Informatics, University College London, London, U.K.
4. NIHR Biomedical Research Centre at South London and Maudsley NHS Foundation Trust and King's College London, London, U.K.

Corresponding author
Richard JB Dobson: richard.j.dobson@kcl.ac.uk

# ABSTRACT

**Objective**

Biomedical documents such as Electronic Health Records (EHRs) contain a large amount of information in an unstructured format. The data in EHRs is a hugely valuable resource documenting clinical narratives and decisions, but whilst the text can be easily understood by human doctors it is challenging to use in research and clinical applications. To uncover the potential of biomedical documents we need to extract and structure the information they contain.

**Materials and methods**

The task at hand is called Named Entity Recognition and Linking (NER+L). The number of entities, ambiguity of words, overlapping and nesting make the biomedical area significantly more difficult than many others.

To overcome these difficulties, we have developed the Medical Concept Annotation Tool (MedCAT), an open-source unsupervised approach to NER+L. MedCAT uses unsupervised machine learning to disambiguate entities. It was validated on MIMIC-III (a freely accessible critical care database) and MedMentions (Biomedical papers annotated with mentions from the Unified Medical Language System (UMLS)).

**Results**

In case of NER+L, the comparison with existing tools shows that MedCAT improves the previous best with only unsupervised learning (F1=0.848 vs 0.691 for disease detection; F1=0.710 vs. 0.222 for general concept detection).

A qualitative analysis of the vector embeddings learnt by MedCAT shows that it captures latent medical knowledge available in EHRs (MIMIC-III).

**Discussion**

Unsupervised learning can improve the performance of large scale entity extraction, but it has some limitations when working with only a couple of entities and a small dataset. In that case options are supervised learning or active learning, both of which are supported in MedCAT via the MedCATtrainer extension.

**Conclusion**

Our approach can detect and link millions of different biomedical concepts with state-of-the-art performance, whilst being lightweight, fast and easy to use.

***K*eywords**: Natural Language Processing · Unsupervised Machine Learning · Electronic Health Records

# BACKGROUND AND SIGNIFICANCE

Electronic Health Records (EHRs) contain invaluable, detailed information about patients' health and the history of care. That information is necessary for optimal clinical decision making. Most of the data in EHRs is stored in an unstructured format (particularly text and images), making it challenging to apply machine learning and statistical modeling. The main objective of this study is to develop a new approach for structuring the text portion of EHRs that is designed to handle the particular complexity of biomedical text, thus enabling multiple downstream applications on top of EHR data.

Advances in NLP such as word embeddings,[1,2] Long-Short Term Memory Networks and transfer learning[3,4] could greatly contribute to the task of extracting information from EHRs. Use-cases such as detection of adverse drug reactions or discovery in disease genomics showcase the possibilities of NLP tools.[5,6] Recently NER models based on Deep Learning (DL), notably Transformers[7] and Long-Short Term Memory Networks have achieved large improvements in accuracy.[8] Both approaches require explicit supervised training, usually involving a human expert to label a certain amount of entities. In the case of medical concepts, very little labelled data is available and very often it can not be made public due to privacy concerns. In addition, medical vocabularies can contain millions of different named entities with overlaps see Figure 1. Considering all this, providing enough training data for DL systems can be extremely challenging.

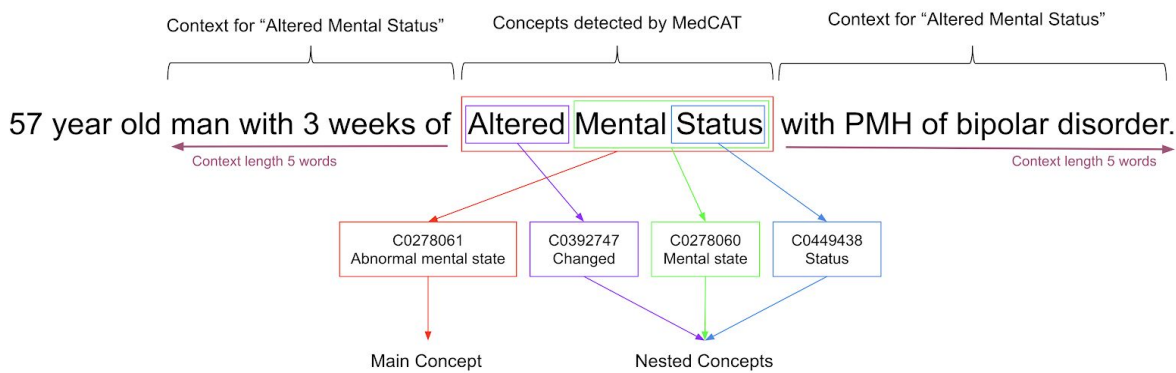

Figure 1. Example of biomedical NER+L with nested entities. Each one of the detected boxes has multiple candidates in the Unified Medical Language System (UMLS). The goal is to detect the entity and annotate it with the most appropriate concept ID, e.g. for the span *Status*, we have three candidates in UMLS, namely *C0449438*, *C1444752*, *C1546481*.

**The main challenges of NER+L in a biomedical context**

Due to the limited availability of training data for supervised learning, in the biomedical domain most NER+L tools use a dictionary-based approach. In this approach we have a vocabulary of all possible terms of interest that could appear in a document (e.g. all UMLS[9] concepts).

This approach allows the detection of concepts without manual annotation, however it poses several challenges such as spelling mistakes, form variability (e.g. *kidney failure* vs *failure of kidneys*), recognition and disambiguation. Recognition is concerned with whether an entity candidate (a portion of text that matches one of the terms in our vocabulary) is a biomedical term or not. This can be illustrated for the UMLS concept with the id *C0030567*, which is linked to the name *Parkinson* (full name *Parkinson's Disease*). Looking at Figure 2 in the second sentence *Parkinson* should be linked to *C0030567* whilst in the first it should not. This problem mainly happens when we have partial or abbreviated names for concepts.

```
Mr Parkinson was admitted to the hospital yesterday because of a...
Patient with severe Parkinson s disease and a history of CVAs...
```

Figure 2. A showcase for the recognition problem that appears for the word *Parkinson*.

Disambiguation is necessary because a name in a biomedical database (e.g. UMLS) can be linked to multiple concepts. Our version of UMLS contains 1.2M different concepts and on average there are two different names assigned to each concept. For example, the concept with id *C0006826* has 16 different assigned names including *cancer, tumour, malignant neoplasm, malignancy and disease.* On average 90% of these names link to more than one concept in UMLS, as a consequence it is impossible to link a detected entity candidate to a biomedical concept based only on the name.

**A brief overview of existing tools for NER+L**

The difficulties mentioned above might be the reason why there are fewer NER+L tools in the biomedical field than in many other areas. Within these, only some can be used independently of the concept type or use-case such as MetaMAP[10], BioYODIE[11], SemEHR[12] and cTAKES[13].

MetaMAP was developed at the National Library of Medicine and is used to map biomedical text to the UMLS Metathesaurus. As seen in Table 1 and Figure 3 some of the main limitations of MetaMAP are: (1) ambiguous concepts and (2) spelling mistakes. BioYODIE is a more recent tool for named entity extraction and linking, addressing some of the problems with MetaMAP. Two major contributions are speed and disambiguation capabilities, but because of the way disambiguation is implemented it either requires a pre-annotated corpus or supervised training. Some of the shortcomings of BioYODIE were addressed with SemEHR. The NER+L system in SemEHR takes the output of BioYODIE and applies manual rules to improve the result. Although writing manual rules can be very time consuming they can produce very good results.[14]

Another NER+L system is cTAKES which builds on existing open-source technologies—the Unstructured Information Management Architecture[15] framework and OpenNLP[16] natural language processing toolkit. The core cTAKES library can not handle any of the 4 NER+L challenges in Table 1 although some can be added through additional plugins.

| **Example** | **Problem** | **cTAKES** | **BioYODIE** | **MetaMAP** | **SemEHR** |
|---|---|---|---|---|---|
| $E_1$ | Disambiguation | - | Heart Rate | - | Hour |
| $E_2$ | Spelling | - | - | - | - |
| $E_3$ | Disambiguation | - | Heart Rate | - | Hour |
| $E_4$ | Form Variability | - | - | Kidney Failure | - |

Table 1. An illustrative example of problems encountered while using the existing NER+L tools. E1-4 in the Example column refer to the text shown in Figure 3.

$$\text{During the night } \overbrace{HR}^{E_1} \text{ was in the 40s-50s and the } \overbrace{pattient}^{E_2} \text{ was given 8mg/} \overbrace{HR}^{E_3} \text{ of ...}$$

$$\text{... } \overbrace{\text{Failure of kidneys}}^{E4} \text{ ...}$$

Figure 3. Two examples of biomedical text used to showcase disambiguation, spelling and resistance to form variability. $E_1$ requires disambiguation and it should be detected as *Heart Rate*, $E_2$ is misspelled and should be detected as *Patient*, $E_3$ is again disambiguation - *Hour*, and finally $E_4$ is another form of the concept *Kidney Failure*.

The amount of training data needed, problems with privacy and the general issues with existing tools such as recognition, disambiguation, spelling and form variability highlight the need for a new approach. Therefore, we have developed a new tool that learns to extract entities from biomedical documents in an unsupervised fashion while remaining lightweight, fast and easy to use. We first prepared the concept database and vocabulary required for NER+L, then we performed unsupervised training and validation on two biomedical sources. Next we have assessed the performance of our tool and compared it with others previously mentioned. We demonstrate the ability of our method to effectively deal with spelling mistakes, form variability and most importantly recognition and disambiguation.

## MATERIALS AND METHODS

The Medical Concept Annotation Tool (MedCAT) learns and extracts entities from biomedical documents. The MedCAT workstream can be split into three phases (Figure 4): (1) Data preparation, (2) Text annotation and (3) Output analysis.

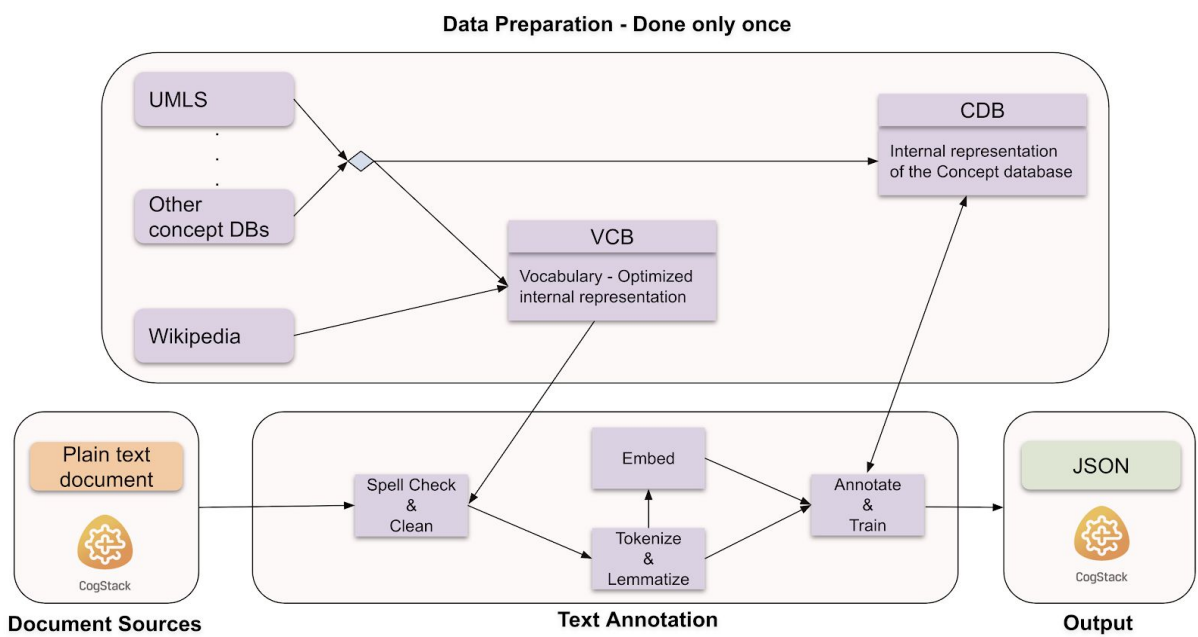

Figure 4: A high-level representation of MedCAT.

**Preparing the concept database and vocabulary required for NER+L**

The data preparation phase is done only once and consists of the following:

1. A Vocabulary (VCB) contains the list of all possible words that can appear in the documents we want to annotate. In our case it is mainly used for spell checking. We have compiled our own VCB by scraping Wikipedia and enriching it with words from UMLS. Only the Wikipedia VCB is made public, the full VCB can be built with scripts provided in the MedCAT repository (https://github.com/CogStack/MedCAT). The scripts require access to the UMLS Metathesaurus (https://www.nlm.nih.gov/research/umls).
2. A Concept Database (CDB) which is built from a biomedical dictionary (e.g. UMLS). Each new concept added to the CDB is represented by an *ID* and *Name*.

Preparing UMLS

UMLS is a set of files and software bringing together many health and biomedical dictionaries and standards, the largest portion of UMLS being the Metathesaurus. It is a large biomedical thesaurus organized by concept, or meaning. Some basic information on UMLS is shown in Table 2. All mentions of UMLS in this document refer to the UMLS Metathesaurus. As UMLS was not built for NER+L tasks, it presents several limitations for our use-case. To overcome these the UMLS was preprocessed as described in Appendix A.

|  | **Original** | **Preprocessed** | **MedMentions** |
|---|---|---|---|
| #Concepts | 3,848,696 | 1,153,689 | 34,550 |
| #Concept Names | 14,608,809 | 2,417,001 | 71,165 |
| #Semantic Types | 127 | 127 | 126 |
| Average links per Concept | 1.8 | 2.18 | 2.55 |
| #Sources | 155 | 80 | 1 |

Table 2. Basic information on the UMLS Metathesaurus before and after preprocessing + the MedMentions subset of UMLS.

**The text annotation pipeline and training**

Once the CDB and VCB have been prepared, we can perform text annotation and run the training procedure. The annotation pipeline starts with cleaning and spell-checking the text, which improves the accuracy of the annotations. We required a fast and lightweight spell checker allowing us to fully control the spelling procedure. Peter Norvig's spell checker implementation (http://www.norvig.com/spell-correct.html) was used (a simple spell-checker using word frequency and the distance between misspelled and correct words to fix mistakes), and optimized for the biomedical domain in the following way:

- A word is spelled against the VCB, but corrected only against the CDB.
- The spelling is never corrected in the case of abbreviations.
- An increase in the word length corresponds to an increase in character correction allowance.
- For efficiency, cache corrections.

Next, the document is tokenized and lemmatized to ensure a broader coverage of all the different forms of a concept name. For this we have used SciSpaCy[17], a tool specifically tuned to these tasks in the biomedical domain. Finally, to detect entity candidates we use a dictionary based approach with a moving expanding window:

1. Given a document $d_1$
2. Set *window_length = 1* and *word_position = 0*
3. There are three possible cases:
    a. The text in the current window is a concept in our CDB, if so mark it and go to step 4
    b. The text is a substring of a longer concept name, if so go to step 4
    c. Otherwise reset *window_length* to 1, increase *word_position* by 1 and repeat step 3
4. Expand the window size by 1 and go back to step 3

Steps 3 and 4 help us solve the problem of overlapping entities shown in Figure 1.

Unsupervised training procedure

To address the difficulties related to disambiguation and recognition we use context similarity. In Figure 3 it is possible to tell whether the word *Parkinson* should be annotated and linked to the concept *C0030567* (*Parkinson's disease*) based on the context. The main idea is to find and annotate mentions of concepts which are unambiguous and from that learn the context. For new documents, when a concept candidate is detected its context is compared to the learned one, if the similarity is high enough the candidate is annotated and linked. The similarity between the context embeddings also serves as a confidence score of the annotation and can be later used for filtering and further analysis. If disambiguation is needed we check the context similarity of each concept that maps to the chosen set of words and take the one with the highest similarity.

The unsupervised training procedure can be defined as follows:

1. Take any corpus of biomedical documents and a biomedical dictionary.
2. For each concept in the CDB ignore all the names that are not unique (ambiguous) or that are known abbreviations.
3. Go through all the documents in the dataset and detect the concept candidates using the approach described earlier. The filtering applied in the previous steps guarantee that the entity candidate can be annotated.
4. For each annotated entity calculate the context embedding $V_{cntx}$.
5. Update the concept embedding $V_{concept}$ with the context embedding $V_{cntx}$.

For the unsupervised training to work one of the names assigned to the biomedical concept must be unique in the biomedical dictionary. The unique name is required as a reference point for training because it is used to learn the context, so that when an ambiguous name appears it can be disambiguated. If we look at the concept with id *C0006826* the unique name is *Malignant Neoplasm*, meaning this name is only assigned to this concept. If we find an entity with this name inside of a biomedical text we can use it for training, because it links only to the *C0006826* concept. Approximately 99% of the concepts in UMLS have at least one unique name.

To represent the context of a concept we use vector embeddings. Given a document $d_1$ where $C_x$ is a detected concept candidate (Equation 1) we need to calculate the context embedding. In other words a vector representation of the context for that concept candidate (Equation 2).

$$d_1 = w_1\ w_2\ \cdots\ \overbrace{w_k\ w_{k+1}}^{Cx}\ \cdots\ w_n \qquad (1)$$

$d_1$   - Example of a document
$w_{1..n}$   - Words/tokens in a document
$C_x$   - Detected concept candidate that matches the words $w_k$ and $w_{k+1}$

$$V_{cntx} = \frac{1}{2s}\Big[\sum_{i=1}^{s} V_{w_{k-i}} + \sum_{i=1}^{s} V_{w_{k+1+i}}\Big] \qquad (2)$$

$V_{cntx}$   - Calculated context embedding
$V_{wk}$   - Word embedding
s   - Words from left and right that are included in the context of a detected concept candidate.
    Typically in MedCAT s is set to 9 for *long* context and 2 for *short* context.

To be able to calculate the context embedding we first require word embeddings which for speed and simplicity were calculated using Word2Vec[1]. Word2Vec was chosen over other more expressive word embeddings like ELMo[18] and BERT[7] for the following reasons: a) ELMo, BERT and similar are not trained with the specific goal that similar words should have similar embeddings; b) Training ELMo is significantly more time consuming and requires a much larger dataset; and c) Hardware requirements for any model using deep learning are much higher than for Word2vec.

We have used the pre-trained Word2vec embeddings made publicly available by Google (https://code.google.com/archive/p/word2vec/) in total 1.4 million words, where each word is represented by a 300 dimensional vector.

Once correct annotations are generated either via unsupervised learning, context embedding $V_{cntx}$ is calculated for each annotated example, and the appropriate $V_{concept}$ is updated using the following formula:

$$sim = max(0, \frac{V_{concept}}{\|V_{concept}\|} \cdot \frac{V_{cntx}}{\|V_{cntx}\|}) \qquad (3)$$

$$lr = \frac{1}{C_{concept}} \qquad (4)$$

$$V_{concept} = V_{concept} + lr \cdot (1 - sim) \cdot V_{cntx} \qquad (5)$$

$C_{concept}$ - Number of times this concept appeared during training
sim   - Similarity between $V_{concept}$ and $V_{cntx}$
lr   - Learning rate

To prevent the context embedding for each concept being dominated by most frequent words, we used negative sampling as explained in.[1] Whenever we update the $V_{concept}$ with $V_{cntx}$ we also generate a negative context by randomly choosing *K* words from the vocabulary consisting of all words in our dataset. Here *K* is equal to *2s* i.e. twice the window size for the context (*s* is the context size from one side of the detected concept, meaning in the positive cycle we will have *s* words from the left and *s* words from the right). The probability of choosing each word and the update function for vector embeddings is defined as

$$P(w_i) = \frac{f(w_i)^{3/4}}{\sum_j^n f(w_j)^{3/4}} \quad (6)$$

$$f(w_i) = \frac{C_{w_i}}{\sum_j^n C_{w_j}} \quad (7)$$

$$V_{ncntx} = \frac{1}{K} \sum_i^K V_{w_i} \quad (8)$$

$$sim = max(0, \frac{V_{concept}}{\|V_{concept}\|} \cdot \frac{V_{ncntx}}{\|V_{ncntx}\|}) \quad (9)$$

$$V_{concept} = V_{concept} - lr \cdot sim \cdot V_{ncntx} \quad (10)$$

n  - Size of the vocabulary
P($w_i$) - Probability of choosing the word $w_i$
K  - Number of randomly chosen words for the negative context
$V_{ncntx}$ - Negative context

**Concept similarity and the co-occurrence matrix**

During the unsupervised training procedure MedCAT calculates context embeddings for concepts in the CDB. These embeddings are required for the disambiguation procedure, but can also be used to calculate similarities between concepts using Equation 2. As shown in the Results section, it allows for a simple way to validate and further explore the tool.

While input text blocks are being annotated by MedCAT, a co-occurrence matrix is also calculated. The text block can include anything from a sentence to all combined medical documents from a patient. The input format is important because the co-occurrence matrix can be updated on the level of the input text block, or whenever two concepts co-occur (meaning it can happen multiple times per text block). The co-occurrence matrix can not only be used to validate the tool, but also for medical use-cases to understand for example which symptoms most frequently co-occur with a certain disease.

**Experimental setup and datasets**

MedCAT was validated on two datasets:

1) MedMentions[19] - The dataset consists of 4,392 titles and abstracts randomly selected from papers released on PubMed in 2016 in the biomedical field, published in the English language, and with both a Title and Abstract. The text was manually annotated for UMLS concepts resulting in 352,496 mentions. During the annotation process, the annotators expanded the UMLS concepts with new names/links. For example, in UMLS, the concept with the ID *C0029456* has two names/links (*Osteoporosis* and *Bone rarefaction*), during the annotation process the annotators expanded that and added *osteoporotic* and *disease*. This means that even though UMLS was used as a base, sometimes the annotators inferred the disease without an explicit mention. Lastly, based on our calculations around 40% of concepts in MedMentions require disambiguation, in other words the detected span of text can be linked to multiple UMLS concepts.

2) MIMIC-III[20] - It is a publicly available dataset developed by the MIT Lab for Computational Physiology, comprising de-identified health data associated with 40,000 critical care patients. MIMIC-III includes data about demographics, vital signs, and laboratory test.

For both datasets UMLS or its subset was used as the biomedical dictionary, and MedCAT was configured as follows:

- The input text was lemmatized and all non alphanumeric characters were skipped.
- Misspelled words were fixed only when 1 change away from the correct word for words under 6 characters, and 2 changes away for words above 6 characters.
- If during unsupervised training a concept appeared < 3 times it was ignored.
- For each concept we calculate *long* and *short* embeddings and take the average of both. The *long* embedding takes into account $s = 9$ words from left and right (as shown in Equation 2). The *short* embedding takes into account $s = 2$ words from left and right. The exact numbers for $s$ were calculated by testing the performance of all possible combinations for $s$ in the range [0, 10]. It was observed that a longer context works better for concepts with the semantic type disease, symptom and similar, while a short context is better for qualitative and quantitative terms.
- The context similarity threshold used for recognition was set to 0.1.

## RESULTS

**Quantitative analysis of the performance of different tools on the MedMentions dataset**

We have tested three configurations of MedCAT on three different versions of the MedMentions dataset:

- Datasets
    - MedMentions Expanded - The full set of annotations in the original format without any changes..

- MedMentions Strict - As mentioned earlier during the annotation process the annotators expanded the base version of UMLS with additional names for concepts. In this dataset we only keep the annotations with names that exist in the base version of UMLS. We are doing this because all the NER+L tools we have tested are built using the base version of UMLS without any extensions.
- MedMentions Strict (Diseases) - Same as the previous dataset, but with the additional restriction to only concepts that are diseases.
- MedCAT configurations (all training is unsupervised)
  - MedCAT U/MI - The default version of UMLS pre-trained on MIMIC-III.
  - MedCAT U/MI/MM - Same as above, i.e. pretrained on MIMIC-III and then fine-tuned on MedMentions.
  - MedCAT MM - The concept database was built using the manual annotations in MedMentions, while the unsupervised training was performed on the text portion of MedMentions.

An annotation by MedCAT is considered correct only if the exact text value was found and the annotation was linked to the correct concept in the CDB. We also tested three related NER+L tools under the exact same conditions as MedCAT: BioYODIE, SemEHR and cTAKES. All tools were used with their default models and parameters, no fine-tuning from our side was performed. The results have been summarized in Table 4.

| *Model \ Dataset* | **MedMentions Expanded** | | | **MedMentions Strict** | | | **MedMentions Strict (Diseases)** | | |
|---|---|---|---|---|---|---|---|---|---|
| | P | R | F1 | P | R | F1 | P | R | F1 |
| SemEHR | 0.169 | 0.276 | 0.222 | 0.279 | 0.284 | 0.281 | 0.495 | 0.888 | 0.691 |
| BioYODIE | 0.161 | 0.276 | 0.218 | 0.270 | 0.284 | 0.277 | 0.490 | 0.885 | 0.687 |
| cTAKES | 0.179 | 0.179 | 0.179 | 0.238 | 0.234 | 0.236 | 0.314 | 0.894 | 0.604 |
| **MedCAT U/MI** | 0.414 | 0.427 | 0.420 | 0.425 | 0.694 | 0.559 | 0.675 | 0.907 | 0.791 |
| **The models below were trained (unsupervised) on the target dataset, on top of that for the last model (MedCAT MM) we used the manual annotations from MedMentions to expand the concept database. As a consequence the models below are not directly comparable with the tools/models above.** | | | | | | | | | |
| **MedCAT U/MM** | 0.434 | 0.525 | 0.479 | 0.422 | 0.858 | 0.640 | 0.720 | 0.977 | 0.848 |
| **MedCAT MM** | 0.613 | 0.807 | 0.710 | - | - | - | - | - | - |
| Support | 352496 | | | 203912 | | | 7615 | | |

Table 3. Metrics for all the tools were calculated in the same way. For each manual annotation we check whether it was detected and linked to the correct Unified Medical Language System (UMLS) concept. The metrics we have used are precision (P), recall (R) and the harmonic mean of precision and recall (F1).

From Table 4 we can see that the best performance for all tools is achieved on the third dataset (disease concepts only). In the case of MedCAT this happens because different diseases appear in a significantly different context, making the embeddings easier to learn and more distinguishable. The table also shows that unsupervised training on the target dataset is helpful and that MedCAT is able to fine-tune the concept embeddings for a specific dataset. For example in the second dataset (MedMentions Strict) the finetuning improves the F1 score from 0.559 to 0.640.

**Qualitative analysis of embeddings on the MIMIC-III dataset**

MedCAT learns vector embeddings for the context in which a concept appears. Consequently, it is possible to perform a qualitative analysis by looking at similarities between concept embeddings, with the expectation that similar concepts have similar embeddings. To calculate the vector similarity between embeddings we use the same formula as in Equation 2. Concept embeddings were trained using the MIMIC-III dataset, the training is unsupervised and done on ~2.4M clinical notes (nursing notes, notes by clinicians, discharge reports etc.).

The training was done on a small one-core server and takes around 30h. In Table 3 and Figure 5 we show that MedCAT learns medically relevant context embeddings and captures medical knowledge including relations between diseases, medications and symptoms.

| *Disease -> Medication* | *Disease -> Procedure* | *Symptom -> Medication* | *Symptom -> Everything* |
|---|---|---|---|
| **Hypertensive disease** | **Cancer** | **Fever** | **Hemorrhage** |
| Metoprolol 50 MG | Chemotherapy | Levofloxacin | Intracranial Hemorrhages |
| Metoprolol 25 MG | Radiosurgery | Vancomycin | Cerebellar hemorrhage |
| Valsartan 320 MG | FOLFOX Regimen | Vancomycin 750 MG | Postoperative Hemorrhage |
| Nadolol 20 MG | Chemotherapy Regimen | Azithromycin | Retroperitoneal Hemorrhage |
| Atenolol 100 MG | Preoperative Therapy | Levofloxacin 750 MG | Amyloid angiopathy |
| Enalapril 10 MG | Anticancer therapy | Dexamethasone | Internal bleeding |
| Oral form diltiazem | Parotidectomy | Lorazepam | Hematoma, Subdural, Chronic |
| niMODipine 30 MG | Resection of ileum | Acetaminophen | Intraparenchymal |

Table 4. Cosine similarity between vector embeddings of concepts. The first row defines the nature of the chosen concept and the targets. We have chosen some of the most frequent concepts and presented foreach the 8 most similar concepts. For example in the second column *cancer* is the chosen concept (disease) and the rows below are the top 8 most similar medications.

$$e_{\text{Kidney Failure}} - e_{\text{Kidney}} + e_{\text{Heart}} = e_{\text{Heart Failure}}$$

$$e_{\text{Metoprolol 25MG}} - e_{\text{Hypertensive Disease}} + e_{\text{Epilepsy}} = e_{\text{Trileptal}}$$

Figure 5. Vector composition of embeddings of concepts. The first example shows that MedCAT learned the relation between organ failure and affected organ, while the second one shows that it learned the relation between diseases and medications.

## DISCUSSION

This paper presents a new tool, MedCAT, that learns to extract entities from biomedical documents in an unsupervised fashion while remaining lightweight, fast and easy to use. We show that MedCAT can effectively deal with spelling mistakes, form variability and most importantly recognition and disambiguation. We performed unsupervised training and validation using two biomedical sources and benchmarked the performance of our tool against existing tools.

One of our main goals was to validate our tool on a publicly available dataset, as this would make tests transparent and facilitate future comparisons. Before MedMentions, to the best of our knowledge, a public dataset of UMLS annotations of this size did not exist. Using MedMentions we have tested the precision and recall on the NER+L task for MedCAT and various other tools. Our results show an improvement on the previous best for NER+L (F1=0.848 vs 0.691 for disease detection; F1=0.710 vs. 0.222 for general concept detection). Furthermore, this improvement in performance is achieved using only unsupervised training.

The particularly large difference in performance for general concept detection could be due to the fact that other tools were trained and fine-tuned to primarily detect diseases and so perform badly on datasets that have a prevalence of qualitative and quantitative concepts. Future studies should further investigate this.

Recent work[21] showed that a deep learning approach (BioBERT+) achieved an F1=0.56 on the full MedMentions dataset whereas MedCAT achieves F1=0.71. We expect this is due to the large number of rare annotations in MedMentions, which is a challenge for a supervised approach (BioBERT+) but can be learned by MedCAT if the concept is not ambiguous. We would expect the performance of BioBERT+ to improve with more training data, however the availability of supervised data is and continues to be a major limitation in biomedical NLP.

To test the quality of the vector embeddings we have trained MedCAT on a medical dataset, namely MIMIC-III. Using a medical dataset was crucial as in such a dataset we expect that similar medical terms have similar embeddings. The vector embeddings learnt by MedCAT show that it captures latent medical knowledge available in EHRs, similar to a recent finding in the context of materials science.[22]

The main advantage of MedCAT is the ability to perform disambiguation and detection based on unsupervised learning alone. Although the focus of the current study is unsupervised learning, MedCAT also supports supervised and active learning models. Both of these enable tailoring of the annotations to a specific use-case. Active learning is especially beneficial as it allows us to fine-tune the annotations once the unsupervised training is done. This provides significant speedups, as most concepts are already trained and we only need to take care of the few concepts that did not receive enough training. MedCAT can interface with any training front-end that supports supervised/active learning, such as MedCATTrainer[23] which was recently built for this purpose.

**Limitations**

While MedCAT allows a significant increase in accuracy compared to other tools, there are still some limitations to be acknowledged. Firstly, when two different concepts appear in a similar context and have the same name (e.g *Mg* - which links to *Magnesium* and *Milligram*, both of which appear in the medications list section of EHRs), the linking procedure may fail. Furthermore if a concept appears in many different contexts, a very large number of examples is needed to learn to detect it correctly. Both situations are rare and usually happen for qualitative or quantitative concepts. Future studies should investigate more expressive vector embeddings for words, or switch to neural networks that calculate embeddings for entire paragraphs.

Secondly, because the training is unsupervised the number of training examples required in the corpus grows with the number of concepts in the concept database. This is due to the increasing probability of names of concepts being ambiguous or appearing in a similar context, which is a particularly significant issue for biomedical concept databases.[24] Based on our experiments we need at least 30 occurrences of a concept in the free text to be able to perform disambiguation (see Appendix B).

Finally, if a concept name in the biomedical dictionary is not unique, we will need to: (1) Manually provide a unique name, or (2) Provide supervised training data. By doing either (1) or (2) we are allowing the context of a concept to be learned, and therefore the synonyms to be recognized correctly. Thus, MedCAT performs at its best when the following assumptions are met: 1) concepts to be detected have unique names and 2) there is sufficient training examples given the size of the concept database.

**Future work**

Although MedCAT showed an improved performance in terms of recognition, disambiguation, spelling and form variability, this tool does not detect negation, temporality, experiencer and similar meta-annotations for a concept (this functionality is provided via the MedCATtrainer extension). While some of the meta-annotations are very simple, others require a deeper understanding of the context. As a result we have decided to focus on supervised deep learning approaches (e.g. RoBERTa[25]) that can embed a larger context - which in turn can significantly improve the accuracy of meta-annotations. More generally, future studies should explore potential options using supervised deep learning approaches. As previously mentioned, a major challenge is that large biomedical databases require extremely large datasets for unsupervised training. A potential solution currently being investigated is to use the relationships between biomedical concepts and with that reduce the amount of training needed.

**CONCLUSION**

In biomedical NER+L we usually have a large biomedical concept database (e.g. UMLS) and a corpus of biomedical documents (e.g. EHRs). Each concept can have many different names and abbreviations, most of the names are ambiguous - meaning they are linked to multiple concepts (Figure 2). MedCAT is based on a simple idea: at least one of the names for each concept is unique and given a large enough corpus that name will be used in a context. As the context is learned from the unique name, when an ambiguous concept is later detected, its context is compared to the learnt context, allowing to find the correct link. By comparing the context similarity we also get confidence scores which allows downstream filtering and further analysis.

The huge volume of medical information that is captured solely in free text creates enormous potential for NLP in healthcare, but the nature of biomedical text presents a number of challenges. Here we show that with state-of-the-art NLP methods, many of these challenges can now be addressed. We have developed MedCAT to enable research and delivery of care applications to leverage the data in clinical text for analysis, without requiring specific infrastructure or hardware. Currently MedCAT is being deployed in a number of hospitals in the UK, as part of the CogStack Platform,[26] to inform clinical decisions with real-time alerting and contribute to patient stratification, recruitment to clinical trials and clinical coding.

## ACKNOWLEDGEMENTS

DMB is funded by a UKRI Innovation Fellowship as part of Health Data Research UK MR/S00310X/1 (https://www.hdruk.ac.uk). RB is funded in part by grant MR/R016372/1 for the King's College London MRC Skills Development Fellowship programme funded by the UK Medical Research Council (MRC, https://mrc.ukri.org) and by grant IS-BRC-1215-20018 for the National Institute for Health Research (NIHR, https://www.nihr.ac.uk) Biomedical Research Centre at South London and Maudsley NHS Foundation Trust and King's College London. RJBD is supported by: 1. Health Data Research UK, which is funded by the UK Medical Research Council, Engineering and Physical Sciences Research Council, Economic and Social Research Council, Department of Health and Social Care (England), Chief Scientist Office of the Scottish Government Health and Social Care Directorates, Health and Social Care Research and Development Division (Welsh Government), Public Health Agency (Northern Ireland), British Heart Foundation and Wellcome Trust. 2. The BigData@Heart Consortium, funded by the Innovative Medicines Initiative-2 Joint Undertaking under grant agreement No. 116074. This Joint Undertaking receives support from the European Union's Horizon 2020 research and innovation programme and EFPIA; it is chaired, by DE Grobbee and SD Anker, partnering with 20 academic and industry partners and ESC. 3. The National Institute for Health Research University College London Hospitals

Biomedical Research Centre. 4. National Institute for Health Research (NIHR) Biomedical Research Centre at South London and Maudsley NHS Foundation Trust and King's College London. AR is supported by 1. Health Data Research UK, which is funded by the UK Medical Research Council, Engineering and Physical Sciences Research Council, Economic and Social Research Council, Department of Health and Social Care (England), Chief Scientist Office of the Scottish Government Health and Social Care Directorates, Health and Social Care Research and Development Division (Welsh Government), Public Health Agency (Northern Ireland), British Heart Foundation and Wellcome Trust; 2. National Institute for Health Research (NIHR) Biomedical Research Centre at South London and Maudsley NHS Foundation Trust and King's College London. AM is supported by Takeda.

This work was supported by the National Institute for Health Research (NIHR) University College London Hospitals (UCLH) Biomedical Research Centre (BRC) Clinical and Research Informatics Unit (CRIU), NIHR Health Informatics Collaborative (HIC), and by awards establishing the Institute of Health Informatics at University College London (UCL).

This work was also supported by Health Data Research UK, which is funded by the UK Medical Research Council, Engineering and Physical Sciences Research Council, Economic and Social Research Council, Department of Health and Social Care (England), Chief Scientist Office of the Scottish Government Health and Social Care Directorates, Health and Social Care Research and Development Division (Welsh Government), Public Health Agency (Northern Ireland), British Heart Foundation and the Wellcome Trust.

This paper represents independent research part funded by the National Institute for Health Research (NIHR) Biomedical Research Centre at South London and Maudsley NHS Foundation Trust and King's College London. The views expressed are those of the author(s) and not necessarily those of the NHS, the NIHR or the Department of Health and Social Care. The funders had no role in study design, data collection and analysis, decision to publish, or preparation of the manuscript.
## Competing interests

The authors declare that no competing interests exist.

# APPENDIX A - Building the UMLS Concept Database

Preparing a concept database (CDB) from UMLS is not a trivial process for several reasons, mainly:

- Some concepts have misspellings in names which can be difficult to detect.
- There are hundreds of different naming conventions for concepts e.g. (1) *Cancer [Process]* (2) *Cancer* (3) *Cancer [SnoMED]*. Consequently, the concept name may or may not include the source dictionary or semantic type. This is very useful for some use-cases, but very disadvantageous for a dictionary based NER.
- Concepts can have abbreviations, and these may not always be labelled as such.
- A significant number of concepts are chemical formulas for which the writing is dependent on the source dictionary.

To overcome the difficulties above, the UMLS was preprocessed in the following way (results can be seen in Table 2):
1. To deal with misspellings, concept names that appear in multiple source dictionaries were favored. For example the concept *C0432423* has the name *cancer* in 10 different source dictionaries, hence that name is set as preferred.
2. Names that have additional information in parentheses were cleaned up, as this is not necessary for the current use-case.
3. For a concept to be denoted as an abbreviation, it needs to be marked as an abbreviation in UMLS, or be of short length and uppercase.
4. To deal with the different writing styles of chemical formulas, only alphanumeric characters were kept. Chemical formulas are much more complicated to deal with, but for our use case this simplification is sufficient.
5. Only the most widely used source dictionaries (Appendix A) were used and all concepts containing more than 6 words were removed (99% diseases/symptoms/drugs are below 7 words).

The UMLS can be downloaded from https://www.nlm.nih.gov/research/umls/index.html, once done it is available in the Rich Release Format (RRF). To make subsetting and filtering easier we import UMLS RRF into a PostgreSQL database (scripts available at https://github.com/w-is-h/umls).

Once the data is in the database we can use the following SQL script to download the CSV files containing all concepts that will form our CDB.

```
# Selecting concepts for all the Ontologies that are used
SELECT DISTINCT umls.mrconso.cui, str, mrconso.sab, mrconso.tty,
tui, sty, def FROM umls.mrconso LEFT OUTER JOIN umls.mrsty ON
umls.mrsty.cui = umls.mrconso.cui LEFT OUTER JOIN umls.mrdef ON
```

```
          umls.mrconso.cui = umls.mrdef.cui
WHERE umls.mrconso.cui IN (SELECT cui FROM umls.mrconso WHERE
lat='ENG' AND
                                    sab='SNOMEDCT_US' OR
                                    sab='HPO' OR
                                    sab='ATC' OR
                                    sab='DRUGBANK' OR
                                    sab='LNC' OR
                                    sab='RXNORM' OR
                                    sab='ICD10PCS' OR
                                    sab='NCI_NICHD' OR
                                    sab='CHV' OR
                                    sab='MTH' OR
                                    sab='OMIM' OR
                                    sab='NCI' OR
                                    sab='MSH' OR
                                    sab='CSP' OR
                                    sab='MEDCIN') AND lat='ENG'
```

**APPENDIX B - MORE ON THE METRICS**

The real scores may be slightly better than what is shown in Table 4., the main reason being that sometimes there are multiple UMLS concepts that can be correctly linked to a name. As an example we can look at the string *patients* in the Figure A from the MedMentions dataset. In the manual annotations *patients* is linked to the concept *C0030705 - Patients*, but MedCAT has linked it to the concept *C0871463 - Surgical Patients*.

```
...is a reliable bedside tool to determine the frailty status of patients
undergoing emergency general surgery. Frail status as...
```

Figure A. An example where the string *patients* can be linked to two concepts, with no clear best. In this case the MedMention annotation is *Patients* while MedCAT annotation is *Surgical Patients*.

**How many examples are enough**

To test the required number of examples to achieve high enough F1 score, we have created a mini-dataset from MedMentions. It contains only two concepts: C0018810 (Heart Rate) and C2985465 (Hazard Ratio). Both concepts have a unique name and the ambiguous abbreviation HR that can link to either one. We have chosen these two concepts, as the abbreviation HR is the most frequent one in MedMentions, given the requirement that it must be ambiguous. Our dataset consists of:

- 60 training examples (30 per concept). In each example the full name of the concept was used, see Figure B.
- 174 test examples, each document contains the ambiguous abbreviation HR, see Figure B.

> Levels of fibrin degradation products (FDP), D-dimer, fibrinogen, the ratio of FDP to fibrinogen, the ratio of D-dimer to fibrinogen, systolic blood pressure, **heart rate**, the Glasgow Coma Scale, pH, base excess, hemoglobin and lactate levels, the pattern of pelvic injury, and injury severity score were measured at hospital admission, and compared between the two groups.
>
> NEAC was assessed by a validated food frequency questionnaire collected at baseline. We categorized the distribution of NEAC into sex - specific quartiles and used multivariable adjusted Cox proportional hazards regression models to estimate **hazard ratios** with 95% confidence intervals (95% CI).
>
> In the overall population radical nephrectomy was not associated with an increased risk of other cause mortality on multivariable analysis compared to nephron sparing surgery (**HR** 0.91, 95% CI 0.6-1.38, p = 0.6).

Figure B. Three samples from the dataset used to test the amount of training samples needed for disambiguation to work. First example is a training case for the concept C0018810, second for C2985465 and third is used to test the disambiguation performance.

We have tested the performance for different sizes of the training set: 1, 5, 10 and 30. If we set the training set size to e.g. 5, we split the full training set into 6 parts (in total the training set has 30 examples per concept), each containing 5 examples per concept. Then we check the performance for each part and report the average over the 6 parts, see Table A.

| Number of examples per concept | F1 on Test |
|---|---|
| 1 | 0.74 |
| 5 | 0.81 |
| 10 | 0.82 |
| 30 | 0.86 |

Table A: Relation between the number of training examples and performance of MedCAT concept disambiguation.